\documentclass[letterpaper, 10 pt, conference]{ieeeconf}  

\IEEEoverridecommandlockouts                              

\newcommand{\RNum}[1]{\uppercase\expandafter{\romannumeral #1\relax}}

\usepackage{graphicx,wrapfig,fullpage,amsmath,hhline,epsfig,verbatim,url,amssymb,multicol,multirow,cite}
\usepackage{times,color,soul} 
\usepackage[left=0.75in,top=0.70in,right=0.75in,bottom=0.80in]{geometry}
\usepackage{amsbsy,latexsym,mathrsfs,mathtools}

\usepackage{mathptmx} 
\DeclareMathAlphabet{\mathcal}{OMS}{cmsy}{m}{n}
\usepackage{bm}
\usepackage{float}
\usepackage{color,soul}
\usepackage{dsfont}
\usepackage{comment}
\usepackage{algorithm}
\usepackage{algorithmic}
\usepackage{psfrag}   

\usepackage{url}
\usepackage{hyperref}
\usepackage{setspace}
\usepackage[table]{xcolor}
\usepackage{paralist}
\usepackage{booktabs}
\usepackage{balance}

\newcommand{\YuanMod}[1]{\textcolor{black}{#1}}

\title{\LARGE \bf
	 Impact-Aware Online Motion Planning for Fully-Actuated Bipedal Robot Walking
}

\author{Yuan Gao, Xingye Da, and Yan Gu
	\thanks{Yuan Gao and Yan Gu are with the Department of Mechanical Engineering, University
of Massachusetts Lowell, Lowell, MA 01854, U.S.A. Emails: {yuan\_gao}@student.uml.edu, {yan\_gu}@uml.edu. Xingye Da is a AI and Robotics Research and Development Engineer in Nvidia Coporation. Email: xda@nvidia.com. }
}

\begin{document}

	\maketitle
	\thispagestyle{empty}
	\pagestyle{empty}

	\begin{abstract}
	The ability to track a general walking path with specific timing is crucial to the operational safety and reliability of bipedal robots for avoiding dynamic obstacles, such as pedestrians, in complex environments. 
	This paper introduces an online, full-body motion planner that generates the desired impact-aware motion for fully-actuated bipedal robotic walking. 
    The main novelty of the proposed planner lies in its capability of producing desired motions in real-time that respect the discrete impact dynamics and the desired impact timing.%
	To derive the proposed planner, a full-order hybrid dynamic model of fully-actuated bipedal robotic walking is presented, including both continuous dynamics and discrete lading impacts.
	Next, the proposed impact-aware online motion planner is introduced.
	Finally, simulation results of a 3-D bipedal robot are provided to confirm the effectiveness of the proposed online impact-aware planner.
	The online planner is capable of generating full-body motion of one walking step within 0.6 second, which is shorter than a typical bipedal walking step.
	\end{abstract}

	\section{INTRODUCTION}
	
	Motion planning of legged robotic locomotion is a challenging problem due to the hybrid, nonlinear, high-dimensional robot dynamics.
	Previously, due to the limited computational power, the motion planning task of legged robotic locomotion was typically conducted offline~\cite{wieber2016modeling}.
	Trajectory optimization was used in offline planning to generate optimal periodic walking patterns~\cite{677298,cabodevila1995energy}.
	Later on, this method was extended to generate non-periodic walking patterns to enable robot navigation in complex, static environments~\cite{kuindersma2016optimization,dai2014whole}.
	
	
	One major limitation of offline planning is that it is not suitable for navigation in dynamic environments (e.g., crowded hallways with moving pedestrians). 
	To help ensure the operational safety and reliability during navigation in dynamic environments, online planning is required.
	For this reason, online planning methods have been extensively investigated in recent years for bipedal robotic walking.
	As bipedal robots typically have high degrees of freedom, online planning methods mainly use reduced-order dynamic models~\cite{kajita20013d,orin2013centroidal} for reducing the computational load \cite{liu2008online,nishiwaki2011simultaneous,nishiwaki2002online}.
	Based on reduced-order dynamics, researchers have utilized Model Predictive Control (MPC) to develop online planning methods for enabling robots to avoid collisions in human-populated areas while maintaining balance~\cite{wieberMPC,bohorquez2016safe,brasseur2015robust}.
	However, these reduced-order dynamic models fail to capture an important, inherent behavior of legged locomotion, which is the swing-foot landing impact. 
	A landing impact occurs when a robot's swing foot strikes the ground, causing a sudden jump in joint velocities as well as an impulsive ground-reaction force.
	Ignoring the impact in planning will result in a significant mismatch between the planned motion and the robot's actual behavior during highly dynamic walking, especially when the leg mass and motor inertia are not negligible. 
	Minimizing the landing impact to zero during motion planning may lead to a ``cautious'' walking style with a limited walking speed, which is also undesirable. 
	To explicitly address landing impacts, the Hybrid Zero Dynamics (HZD) framework~\cite{westervelt2007feedback,grizzle2001asymptotically,Hybrid_zero_dynamics_of_planar_biped_walkers} has been formulated based on full-order dynamic modeling of both continuous and discrete behaviors involved in walking.
	Although the HZD approach mainly focuses on offline planing and periodic walking pattern generation, researchers have incorporated control barrier functions~\cite{nguyen20163d}, deep learning~\cite{siravuru2017deep}, and gait library~\cite{nguyen2016dynamic,da2017supervised} into the framework for realizing online impact-aware planning. 

	Previously, we have theoretically developed the global-position tracking planning and control framework, which explicitly addresses the landing impact dynamics and realizes provably accurate tracking of non-periodic time trajectories on a planar fully-actuated robot ~\cite{gu2016bipedal,gu2018exponential}.
    Later on, we have extended our framework to a three-dimensional (3-D) fully-actuated robot~\cite{gu2018straight,yuan_2018} as well as a planar multi-domain robot~\cite{yuan2019DSCC}.
	There are two main limitations of our previous works: a) the desired motions are generated through offline trajectory optimization and b) the desired walking paths are straight lines.
	Therefore, this paper will incorporate online motion planning into our global-position tracking framework to enable dynamic, stable walking along general-shaped walking paths.

	Unlike following a straight-line walking path, walking along a 3-D general-shaped walking path naturally involves complex footstep sequences.
	In this work, we assume that such a footstep sequence is provided by a higher-level planner, including the position and orientation of each footstep as well as the desired timestamps of foot placement.
	If a robot can reliably track these desired footstep sequences and the corresponding timestamps, we consider that the robot is capable of reliably tracking a 3-D general-shaped walking path with the desired timing.
	The major challenge of the proposed online planning is to generate impact-aware full-body motions in real-time while respecting impact dynamics and other computationally heavy constraints. 
	The impact-awareness constraint is computationally expensive due to the highly nonlinear impact dynamics. 
    For straight-line periodic walking, it takes approximately 8 minutes to generate only one single walking step with the impact-awareness condition satisfied in our previous work~\cite{yuan_2018}. 
	Motivated by the current research needs, this paper proposes an online full-body motion planner that generates impact-aware nonperiodic motions in real-time.
	
	
	This paper has two major contributions.
	The first contribution is the development of an online impact-aware planner that generates desired full-body motion profile. 
	The second contribution is the introduction of a novel method, which is termed as keyframe posture library, to reduce the computational load for realizing efficient planning. 
	
	The paper is organized as follows. 
	The hybrid, floating-based, full-order model of biped robotic walking is presented in Section~\ref{Section-Dyanmics}. 
	In Section~\ref{Section-planner}, the keyframe posture library method is introduced, along with the formulation of a set of optimization problems for creating the proposed motion planner.
	Simulation results are discussed in Section~\ref{Section-Simulation}.

\section{HYBRID FLOATING-BASED DYNAMICS OF BIPEDAL ROBOTIC WALKING}
\label{Section-Dyanmics}

This section presents a full-order model of hybrid bipedal walking dynamics.
As walking inherently involves both continuous dynamics and discrete behaviors, it is natural to model bipedal walking as a hybrid dynamical system.
The following assumptions are considered in this study:
\begin{itemize}
    \item The walking surface is flat and horizontal.
    \item During continuous phases, the support foot remains a static, full contact with the walking surface.
    \item The impact is modeled as a rigid-body contact, which occurs within an infinitesimal period of time~\cite{grizzle2001asymptotically}.
\end{itemize}
Based on these assumptions, the robot is fully actuated during continuous phases.

\begin{figure}[h]
    \centering
    \includegraphics[width=0.95\linewidth]{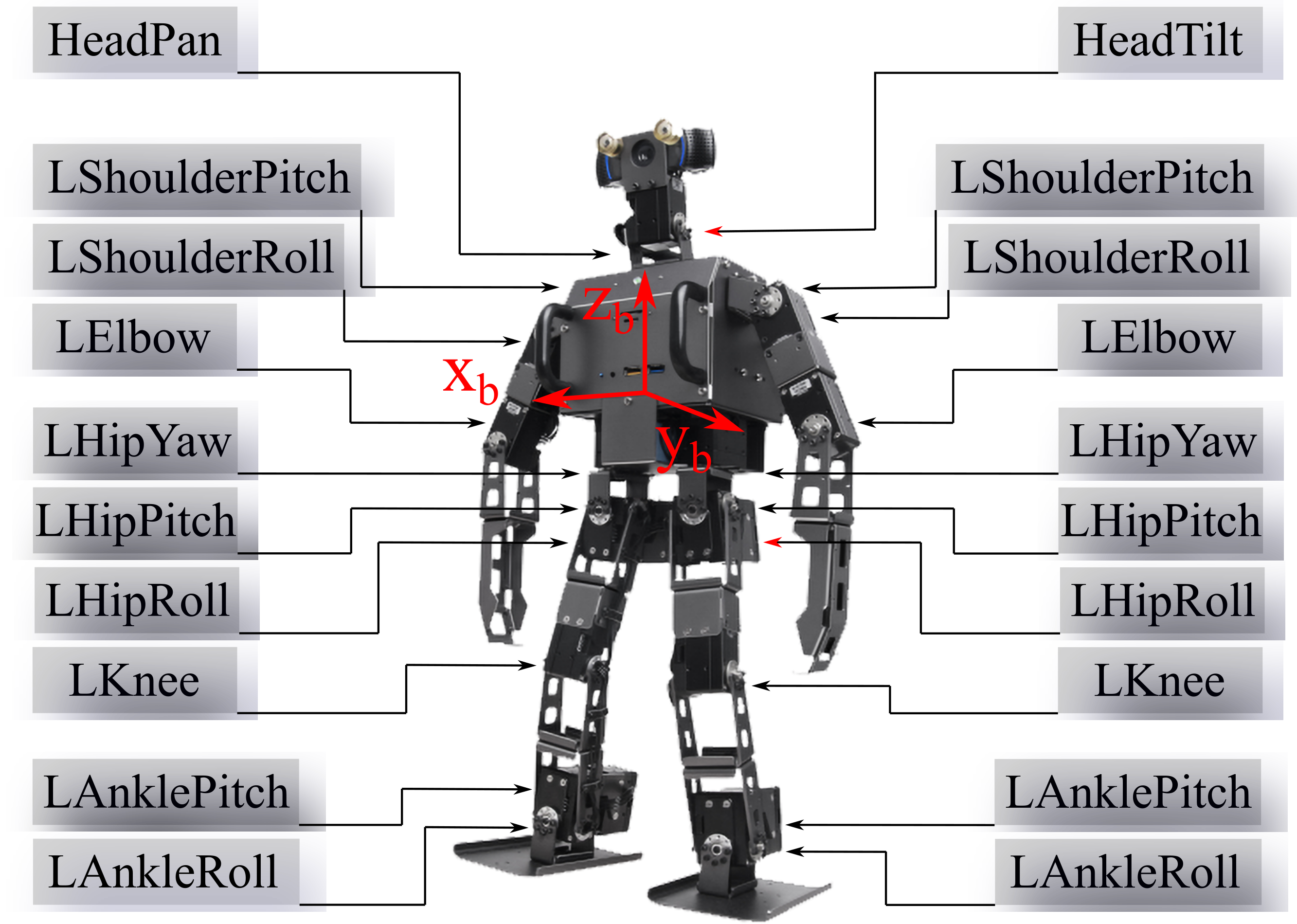}
    \caption{An illustration of the revolute joints of a ROBOTIS-OP3 bipedal humanoid robot.
    The coordinate system of the robot's floating base is located at the center of the chest.}
    \label{fig:ROBOTIS-op3}
\end{figure}

The generalized coordinates of the floating-base bipedal robot can be expressed as 
\begin{equation}
    \begin{bmatrix}
    \mathbf{p}_b^T, \boldsymbol{\gamma}_b^T,q_1,...,q_{n} \end{bmatrix} ^T\in \mathcal{Q},
\end{equation}
where $\mathcal{Q} \subset \mathbb{R}^{n+6}$ is the configuration space, $\mathbf{p}_b:=[x_b,y_b,z_b]^T \subset \mathbb{R}^{3}$ represents the position of the floating-base with respect to (w.r.t) the world coordinate frame,
$\boldsymbol{\gamma}_b:=[\phi_b,\theta_b,\psi_b]^T$ represents the pitch, roll, and yaw angles of the floating base w.r.t. the world coordinate frame,
and $q_1,...,q_n$ represent the robot's joint angles.
The robot model used for simulation validation is ROBOTIS-OP3~\cite{robotisWeb} (Fig.~\ref{fig:ROBOTIS-op3}), which has $20$ (i.e., $n=20$) independent joints. 

\subsection{Continuous Dynamics}

The continuous-phase equation of motion is obtained through Lagrange's method:
\begin{equation}
    \label{equ:dynamics}
	\mathbf{M(q)\ddot{q}+c(q,\dot{q}) = Bu}+\mathbf{J}^T\mathbf{F},
\end{equation}
where $\mathbf{M}(\mathbf{q}): \mathcal{Q} \rightarrow \mathbb{R}^{(n+6) \times (n+6)}$ is the inertia matrix,
$\mathbf{\mathbf{c}(\mathbf{q},\dot{\mathbf{q}})}:  \mathcal{TQ} \rightarrow \mathbb{R}^{(n+6) \times 1}$ is the sum of Coriolis, centrifugal, and gravitational terms,
$\mathbf{B}\subset \mathbb{R}^{(n+6) \times m}$ $(m=20)$ is a constant matrix,
$\mathbf{u}\subset \mathbb{R}^m$ is the input vector,
$\mathbf{F}\subset \mathbb{R}^6$ is the vector of the generalized external force caused by the contact between the support foot and the ground,
and $\mathbf{J(q)}: \mathcal{Q}\rightarrow \mathbb{R}^{6 \times (n+6)}$ is the corresponding Jacobian matrix.

The holonomic constraints that the robot is subject to can be expressed as:
\begin{equation}
    \label{equ-holonomic constraint}
    \mathbf{J\ddot{q}+\dot{J}\dot{q}=0}.
\end{equation}
Combining Eqs.~\eqref{equ:dynamics} and ~\eqref{equ-holonomic constraint} yields the complete continuous dynamics, which can be expressed as:
\begin{equation}
    \label{equ:complete dynamics}
	\mathbf{M(q)\ddot{q}+\bar{c}(q,\dot{q}) = \bar{B}u},
\end{equation}
where $\mathbf{\bar c(q,\dot{q})}:=\mathbf{c-J^T(J M^{-1} J^T)^{-1}(J M^{-1}c-\dot{J} \dot{q})}$ and $\mathbf{\bar B}(\mathbf{q}):=\mathbf{B-J^T(J M^{-1} J^T)^{-1}J M^{-1}B}$. 
Details of the derivation can be found in~\cite{yuan_2018}.

\subsection{Switching Surface}

The switching surface that represents a foot-landing event can be defined as:
\begin{equation}
    \label{equ:switching}
	S_q(\mathbf{q,\dot{q}}):=\{(\mathbf{q,\dot{q}})\in \mathcal{T Q} : z_{sw}(\mathbf{q})=0,\dot{z}_{sw}(\mathbf{q},\dot{\mathbf{q}})<0\},
\end{equation}
where $z_{sw}:\mathcal{Q} \rightarrow \mathbb{R}$ represents the swing-foot height above the ground and $\dot{\mathbf{q}}<0$ indicates that the swing-foot is moving toward the ground.

\subsection{Discrete Dynamics}

Upon a swing-foot landing, an instantaneous rigid-body impact occurs. This impact does not cause discontinuities in joint positions, but joint velocities will experience a sudden jump.
The joint velocities right after an impact can be described as:
\begin{equation}
    \label{equ:reset map}
    \mathbf{\dot{q}}^+=\mathbf{R}_{\dot{q}}(\mathbf{q}^-)\dot{\mathbf{q}}^{-},
\end{equation}
where $\dot{\mathbf{q}}^-$ and $\dot{\mathbf{q}}^+$ represent the the joint velocities right before and after the impact, respectively.
Here,  $\mathbf{R}_{\dot{q}}: \mathcal{Q} \rightarrow \mathbb{R}^{(n+6)\times (n+6)}$ can be obtained from solving the following equation~\cite{grizzle2001asymptotically}: 
	\begin{equation*} 
		\begin{bmatrix}
			\mathbf{M(q)}& \mathbf{-J}^T(\mathbf{q})\\ \mathbf{J}^T (\mathbf{q}) & \mathbf{0}_{6 \times 6} 
		\end{bmatrix}
		\begin{bmatrix}
		\mathbf{\dot{q}^+} \\ \mathbf{\delta F} \end{bmatrix}
		=\begin{bmatrix}
		\mathbf{M(q)\dot{q}^-} \\ \mathbf{0}_{6 \times 1}
		\end{bmatrix},
	\end{equation*}
	where $\delta \mathbf{F}$ is the impulsive ground-reaction force and $\mathbf{0}_{6 \times 6}$ is a $6 \times 6$ zero matrix.

\section{ONLINE IMPACT-AWARE FULL-BODY MOTION PLANNING}
\label{Section-planner}

	This section introduces our proposed impact-aware online motion planner.
	In this study, it is assumed that the desired footstep sequence has been provided by a higher-level planner, including the position and orientation of each footstep as well as the desired timestamps of foot placement.
	The focus of this study is then to generate the full-body motion profile given the desired footstep sequence.
	Specifically, to generate the desired motion profile for one walking step, the input of the planner is the two adjacent footsteps with the given timestamps, and the output of the planner is a set of isolated way-points (i.e., the desired motion profile).
	These way-points can be interpolated to generate a continuous trajectory, which is not the focus of this study and will be addressed in our future work.
	
	
	Planning an impact-aware full-body motion profile is computationally expensive due to the highly complex constraints inherently associated with walking motions.
	These constraints include impact-awareness constraint, which requires that the planned motion should respect the discrete landing-impact dynamics, as well as continuous-phase feasibility constraints. 
	It is necessary to meet these constraints in motion planning because they guarantee the feasibility of the planned motion.
	
\begin{figure}[h]
	    \centering
    \includegraphics[width=1\linewidth]{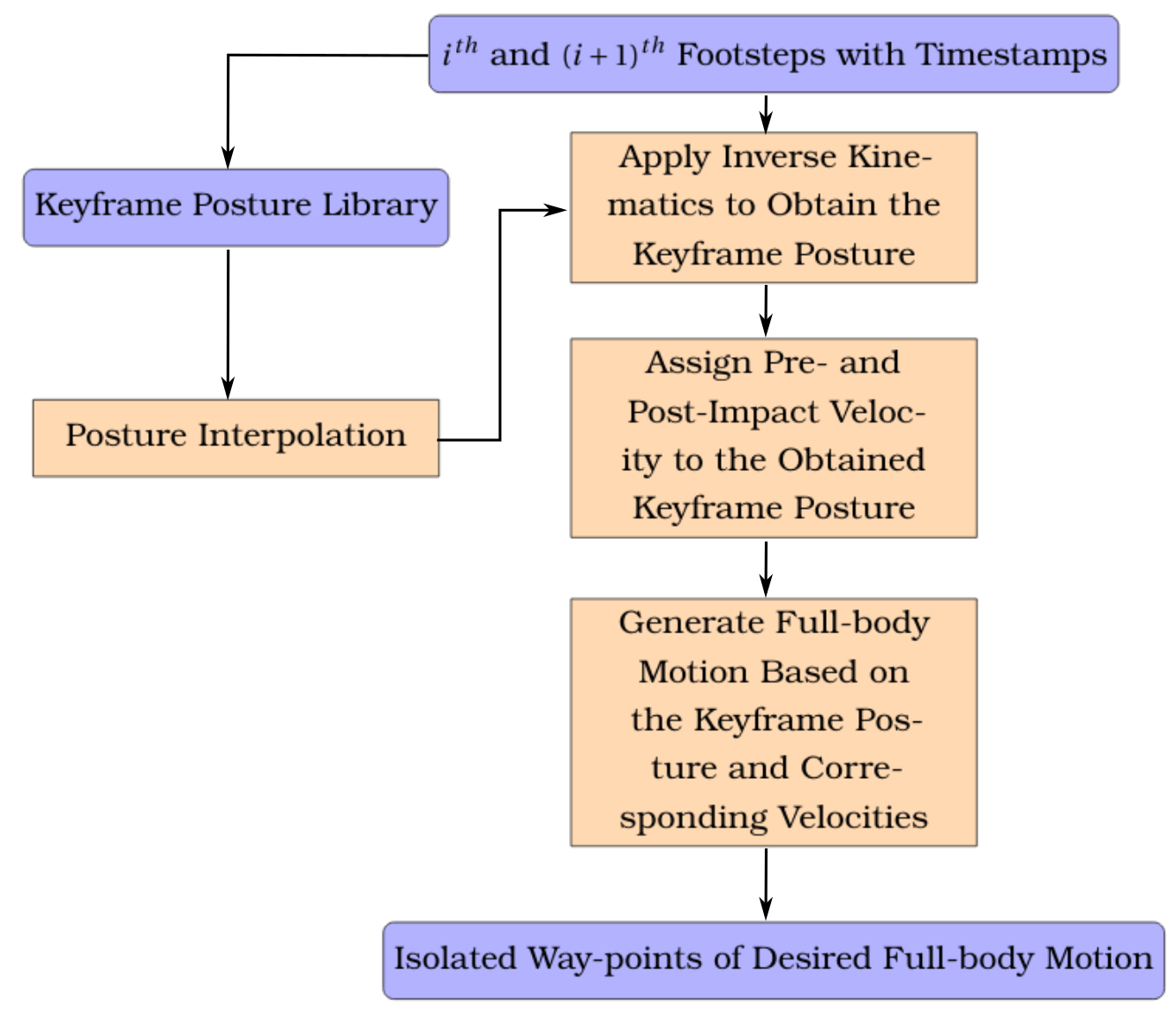}
    \caption{Flowchart of the proposed planner.
    The orange boxes indicate the planning steps, and the blue boxes indicate the data involved in the online planning. 
    The Keyframe Posture Library is highlighted with the blue box as it is pre-computed data.}
    \label{fig:flowchart}
	\end{figure}

	To alleviate the computational load for enabling online planning, we first decompose the complete planning task into four subtasks (Fig.~\ref{fig:flowchart}) such that the impact-aware and the continuous-phase feasibility constraints can be handled separately.
	These four subtasks include: a) posture interpolation, b) computing keyframe postures through inverse kinematics, c) pre- and post-impact velocity assignment to keyframe postures, and d) continuous-phase motion generation.
	%
	
	To further mitigate the computational load of computing keyframe posture, an offline keyframe posture library is introduced and constructed.
	To speed up the planning for continuous-phase motion generation, reduced-order dynamic models are utilized.
	By decomposing the planning task into smaller elements and utilizing both pre-computed results and reduced-order models, 
	our planner is able to generate the impact-aware full-body motion profile of one walking step within 0.6 second, which is typically less than the duration of one walking step. 
	By planning the motion one walking step ahead, the robot is able to move constantly without pausing.

In the following subsections, we will introduce the definition of the keyframe posture, the construction of a keyframe posture library, posture interpolation, and continuous-phase motion generation.
	
	\subsection{Keyframe Posture and Keyframe Posture Library}

\begin{figure}[h]
	    \centering
    \includegraphics[width=0.8\linewidth]{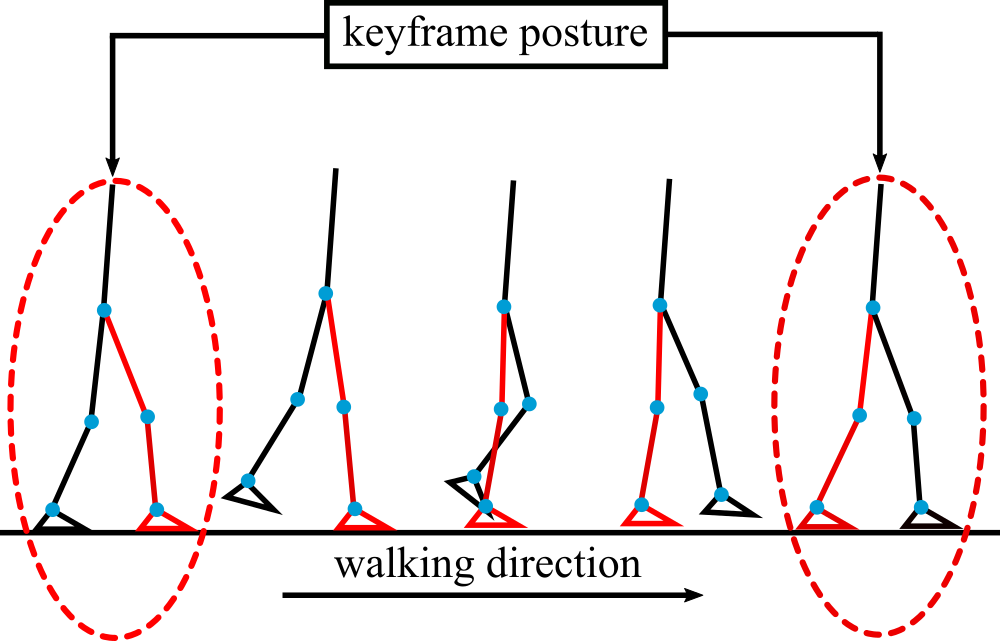}
    \caption{An illustration of keyframe postures during walking.}
    \label{fig:keyframe posture}
	\end{figure}

	Here, we introduce the term, a keyframe posture, to define a pre-computed, kinematically feasible configuration of a walking robot at a swing-foot landing moment
	(Fig. \ref{fig:keyframe posture}).
	We denote the keyframe posture as $\mathbf{q}^k$, where the superscirpt ``$k$'' stands for ``keyframe".
	%
	With offline computing, we can construct a collection of keyframe postures that correspond to a set of relative displacements and orientations between two support feet.
	This collection of the keyframe postures is called \textit{keyframe posture library} ($KPL$):
	\begin{equation} \label{equ:KPL}
	    KPL = \{^{i}\mathbf{q}^{k}|i\in \mathbb{Z}^+,i\le m \},
	\end{equation}
	where $^i\mathbf{q}^k\in \mathcal{Q}$ is the $i^{th}$ keyframe posture within the library, $\mathbb{Z}^+$ is the set of positive integers, and the $m$ is the total number of postures stored in this library.

	The keyframe posture library is an important component of our proposed planner, which provides a feasible initial guess for solving the inverse kinematics associated with meeting the impact-aware constraint.
	The details are discussed next.
    
    \subsection{Posture Interpolation}
    
    Let $\Gamma$ denote the desired footstep sequence provided by a higher-level planner, which is mathematically expressed as 
	\begin{equation}\label{equ:footstep suquence with timestamp}
    \Gamma = \{^i\mathbf{\gamma} : i \in \mathbb{Z}^+, ^i\mathbf{\gamma} \subset \mathbb{R}^4 \},
    \end{equation}
    where $^i\mathbf{\gamma}:=  [^ix^\gamma,^iy^\gamma,^i\phi^\gamma,^i\tau^\gamma]^T$ represents the pose and timestamp of the $i^{th}$ footstep in the given sequence. 
    $^ix^\gamma$, $^iy^\gamma$, and $^i\phi^\gamma$ are the $x$, $y$ and yaw angle of the $i^{th}$ footstep with respect to the world coordinate frame.
    $^i\tau^\gamma$ is the timestamp of the footstep, indicating the desired moment for the robot to step onto that footstep. 
    As this work addresses flat terrain and a full contact between the foot and the ground is assumed, the height, roll and pitch angle of the support foot are all $0$. 
    Thus, we only need to specify $x$, $y$ and yaw angle of the support foot. 
    
    Given two adjacent desired footsteps from $\Gamma$,
    the objective of this step is to obtain the desired posture $^i\mathbf{q}^* \in \mathbb{R}^{n+6}$ that are compatible with the two footsteps.
    One can perform inverse kinematics (IK) to solve this problem. 
    However, this IK problem is nonlinear, non-square and has infinitely many solutions, among which many can be infeasible.
    In order to get a feasible solution efficiently, we can exploit the pre-computed feasible postures in the proposed $KPL$ as explained next:
    \begin{itemize}
        \item Given two adjacent footsteps, search for the two postures within the $KPL$ that correspond to two footsteps closest to the given pair in terms of the relative displacement and orientation. 
        Let these two postures be $^n\mathbf{q}^k$ and $^m\mathbf{q}^k$.
        \item Compute the initial guess of the inverse kinematics by $\mathbf{q}_0=(^n\mathbf{q}^k + ^m\mathbf{q}^k)/2$.
        \item Perform the inverse kinematics to obtain the feasible posture $^i\mathbf{q}^*$.
    \end{itemize}
    As the initial guess of the inverse kinematics (i.e., $\mathbf{q}_0$) is obtained from the $KPL$, one can expect that the solution will be highly likely feasible, which helps to guarantee the reliability of the proposed planner. 
    
    \subsection{Velocity Assignment to Keyframe Postures}
    
	The key novelty of the proposed online planner lies in its capability of generating full-body motions that respect the impact. 
	To satisfy the impact-awareness condition, we assign the pre- and post-impact velocities to each keyframe posture, which is explained next.
	
	Based on the displacement and the timestamp differences between two adjacent footsteps, the average velocity between $i^{th}$ and $(i+1)^{th}$ footsteps ${}^i_{i+1}\mathbf{v}$ can be simply computed as
	\begin{equation}
	    {}^i_{i+1}\mathbf{v} 
	    =
	    \begin{bmatrix}
	    \frac{^{i+1}x^\gamma-^{i}x^\gamma}{^{i+1}\tau^{\gamma}-^{i}\tau^{\gamma}} \\
	    \\
	    \frac{^{i+1}y^\gamma-^{i}y^\gamma}{^{i+1}\tau^{\gamma}-^{i}\tau^{\gamma}} \\
	    \\
	    \frac{^{i+1}\phi^\gamma-^{i}\phi^\gamma}{^{i+1}\tau^{\gamma}-^{i}\tau^{\gamma}}
	\end{bmatrix}.
	\end{equation}
	Then, an optimization problem is formulated to solve for the velocities assigned to the keyframe postures.
	It is important to note that any keyframe posture $^i\mathbf{q}^*$ are associated with two velocities, pre-impact velocity $^i\dot{\mathbf{q}}^-$ and post-impact velocity $^{i+1}\dot{\mathbf{q}}^+$. 
	The optimization problem of solving for the velocities assigned to the posture $^i\mathbf{q}^*$ can be formulated as follows:
	\begin{equation}\label{equ:velocity assignment}
        \begin{aligned}
        & \underset{^i\mathbf{\dot{q}}^-,^{i+1}\mathbf{\dot{q}}^+}{\text{min}}
        & & \mathbf{V}^{-T}\mathbf{Q}\mathbf{V}^- +\mathbf{V}^{+T}\mathbf{P}\mathbf{V}^+\\
        & \text{s.t.} & &  \mathbf{J}_1 ~{}^{i+1}\dot{\mathbf{q}}^+ = \mathbf{0_{6 \times 1}} &~\text{(C11-1)}\\
        & & &  \mathbf{J}_2 ~{}^{i}\dot{\mathbf{q}}^- = \mathbf{0_{6 \times 1}} &~\text{(C11-2)}\\
        & & &  \dot{z}_{sw}(\mathbf{q,{}^i\dot{q}^-}) <0 &~\text{(C11-3)} \\
        & & & [{}^{i+1}\dot{q}_1^+ ,{}^{i+1}\dot{q}_2^+][cos{}^i\phi^\gamma,sin{}^i\phi^\gamma ]^T>0 &~\text{(C11-4)}\\
         & & {}^{i+1}&\mathbf{\dot{q}^+=R_{\dot{q}}(q^-)}~{}^{i}\dot{\mathbf{q}}^{-} &~\text{(C11-5)}
        \end{aligned}
    \end{equation}
    where
    $$\mathbf{V}^-=[\dot{x}^-_b, \dot{y}^-_b, \dot{\psi}^-_b]^T - _{i+1}^{i}\mathbf{v}$$ 
and
    $$\mathbf{V}^+=[\dot{x}^+_b, \dot{y}^+_b, \dot{\psi}^+_6]^T - _{i+1}^{i}\mathbf{v},$$
    $\dot{x}_b$ and $\dot{y}_b$ are the velocities of the robot's base in $x$- and $y$-directions w.r.t. the world coordinate frame,
    and $\dot{\psi}_b$ is the yaw rate of the base w.r.t. the world coordinate frame.
    $\mathbf{Q}\in \mathbb{R}^{3 \times 3}$ and $\mathbf{P}\in \mathbb{R}^{3 \times 3}$ are any positive definite matrices\YuanMod.
    $\mathbf{J}_1$ and $\mathbf{J}_2$ are the contact Jacobian matrices, which are used to enforce the holonomic constraint at the contact points.
    This cost function ensures that the obtained pre- and post-impact velocities are close to the average speed during one step, which helps to prevent dramatic changes in the desired velocity during one step.  
    
    The constraints are explained as follows:
    \begin{itemize}
    \item The constraint $\text{(C11-1)}$ requires that right after the impact, the leading foot should become static on the ground.
    \item The constraint $\text{(C11-2)}$ requires that right before the impact, the trailing foot should be static on the ground.
    \item The constraint $\text{(C11-3)}$ requires that right before the impact, the leading foot should move toward the ground.
    \item The constraint $\text{(C11-4)}$ requires that right after the impact, the velocity of the robot's base should not move backward.
    \item The constraint $\text{(C11-5)}$ is the full-order dynamic relationship between the pre-impact and post-impact velocities.
    \end{itemize}
    
    This optimization problem can be solved efficiently by many optimization toolboxes, such as MOSEK~\cite{mosek} and fmincon~\cite{Matlab_link}.
    It is important to note that the assigned pre- and post-impact velocities automatically satisfy the impact-awareness condition. 
    \subsection{Full-Body Motion Generation}
    
    This subsection presents the last step of our proposed online planning method, which is continuous-phase motion generation.
    To enable online planning, it is reasonable to use reduced-order dynamic model for continuous phases because it significantly reduces the computational cost.
    Centroidal dynamics~\cite{orin2013centroidal} is a well-studied reduced-order dynamic model, which establishes the relationship between the external force/torque and the full-body angular momentum.
    Besides walking, this approach has been used to generate impressive, complex motions, such as jumping and monkey bar~\cite{dai2014whole}.
    These complex motions include significant upper-body rotational motions, during which the centroidal momentum cannot be ignored.
    However, during regular walking, the upper-body motion is trivial, thus the centroidal momentum may be ignored~\cite{brasseur2015robust}. 
    Researchers have previously used Center of Mass (CoM) dynamics to successfully generate continuous-phase walking motions~\cite{wieber2016modeling}.
    The CoM dynamics can be expressed as:
    \vspace{+0.1 in }
    \begin{equation}\label{equ:COM dynamics}
        m\ddot{\mathbf{r}} = \sum_{i=1}^{j} \mathbf{F}_i +m\mathbf{g},
             \vspace{+0.1 in}
    \end{equation}
where $m$ is the robot's total mass, $r \in \mathbb{R}^3$ is the CoM position w.r.t. the world coordinate frame,
$i$ is the $i^{th}$ contact point,
$j$ is the total number of contact points, 
and $\mathbf{F}_i\in \mathbb{R}^3$ is the ground-reaction force applied at the $i^{th}$ contact point. 


To compute the desired continuous-phase motion, Eq.~\eqref{equ:COM dynamics} is converted into difference equations to formulate the nonlinear optimization problem.
Also, the full-order kinematics is considered in the optimization. 
In this case, we sample $K$ points during each step.
The optimization problem is solved for each step in real-time to obtain the desired continuous-phase motion that are dynamically feasible. 
Without loss of generality, 
we use the following cost function for our nonlinear optimization during the $i^{th}$ step: 
    \vspace{+0.1 in }
	\begin{equation} 
	\label{equ:trajectory generation}
        \begin{aligned}
        & \underset{\begin{aligned}^i&\mathbf{q}[k],^{i}\mathbf{\dot{q}}[k],{}^idt[k],\\{}^i&\mathbf{r}[k],{}^i\mathbf{\dot{r}}[k],^i\mathbf{\ddot{r}}[k],\\^{i}&\mathbf{F_j}[k]\end{aligned}}{\text{min}}
        & & {\begin{aligned}\sum_{i=1}^{K} (\|{}^i\mathbf{q}[k]-\mathbf{q}_{norm}[k]\|+\|^{i}\mathbf{ \dot{q}}[k]\|+ \|{}^i\ddot{\mathbf{r}}[k]\|\\+\sum_{i=1}^{j}\|{}^i\mathbf{F}_j[k]\|)\end{aligned}},
        \end{aligned}
            \vspace{+0.1 in }
    \end{equation}
     where ${}^i \star[k]$ indicates the value of $\star$ at $k^{th}$ point during the $i^{th}$ step.
     $\mathbf{q}_{norm}$ is a single pre-computed nominal walking trajectory.
     The same single $\mathbf{q}_{norm}$ is used in the cost function of any $i^{th}$ step, and
     the sole purpose of using $\mathbf{q}_{norm}$ is to help ensure that the generated continuous-phase motion will not have drastically varying joint positions~\cite{dai2014whole}.
    
    The constraints for this optimization include: 
    \begin{itemize}
        \item Dynamic constraint:
         \begin{equation} \tag{C12-1}
        m{}^i\ddot{\mathbf{r}}[k] = \sum_{i=1}^{j} {}^i\mathbf{F}_j[k] +m\mathbf{g}
        \label{equ:dynamic constraint}
    \end{equation}
        \item Kinematic constraint:
        \begin{equation} \tag{C12-2}
            {}^i\mathbf{r}[k]={}^i\mathbf{r}({}^i\mathbf{q}[k]) 
            \label{equ:kinematic constraint}
        \end{equation}
        \item Step duration constraint:
        \begin{equation} \tag{C12-3}
            \sum_{i=1}^{K}{}^idt[k] = {}^{i+1}\tau^\gamma-{}^{i}\tau^\gamma
             \label{equ:step duration constraint}
        \end{equation}
        \item Holonomic constraint:
        \begin{equation} \tag{C12-5}
            \mathbf{J}({}^i\mathbf{q}[k]){}^i\mathbf{\dot{q}}[k]=\mathbf{0_{6 \times 1}}
            \label{equ:holonomic constraint}
        \end{equation}
        \item Keyframe posture constraint:
        \begin{equation} \tag{C12-4}
        \begin{aligned}
            {}^i\mathbf{q}[1] &= {}^i\mathbf{q}^*\\ {}^i\mathbf{q}[K]&={}^{i+1}\mathbf{q}^*\\ {}^i\mathbf{\dot{q}}[1]&={}^i\mathbf{\dot{q}}^{+}\\ {}^{i}\mathbf{\dot{q}}[K]&={}^i\mathbf{\dot{q}}^-
        \end{aligned}
        \label{equ:keyframe posture constraint}
        \end{equation}
        \item Derivative approximation constraint:
        \begin{equation}\tag{C12-6}
        \begin{aligned}
            \dot{\mathbf{r}}[k] = \tfrac{\mathbf{r}[k+1]-\mathbf{r}[k]}{dt[k]}\\
             \ddot{\mathbf{r}}[k] = \tfrac{\mathbf{\dot{r}}[k+1]-\mathbf{\dot{r}}[k]}{dt[k]}\\
              \dot{\mathbf{q}}[k] = \tfrac{\mathbf{{q}}[k+1]-\mathbf{{q}}[k]}{dt[k]}
        \end{aligned}
            \label{equ:derivative approximation constraint1}
        \end{equation}
    \end{itemize}

   The constraints are explained next:
   \begin{itemize}
       \item     The dynamic constraint~\eqref{equ:dynamic constraint} requires that the planned motion satisfies Newton's law.
    \item The kinematic constraint~\eqref{equ:kinematic constraint} indicates the kinematic relationship between the CoM and the configuration of the robot.
    \item The keyframe posture constraint~\eqref{equ:keyframe posture constraint} ensures that the planed motion at the first and the last points ($K^{th}$) equals to the corresponding keyframe postures and velocities.
    \item The holonomic constraint~\eqref{equ:holonomic constraint} ensures that the support foot is static on the ground during the step.
    \item The derivative approximation constraint~\eqref{equ:derivative approximation constraint1} is the finite difference method to compute the derivative in numerical computation.
   \end{itemize}

    
\section{Global-Position Tracking CONTROL}  
\label{Section-Control}
This section introduces a global-position tracking control law as an extension of our previous work~\cite{yuan_2018} from straight-line to general-shaped path tracking.
This controller will utilized in simulation to help validate our proposed online motion planner. 

\subsection{Trajectory Tracking Errors}

Let $\mathbf{h}_c(\mathbf{q}): Q \rightarrow Q_c \subset \mathbb{R}^{n}$ denote the variables of interest.
Let $\mathbf{h}_d(t):\mathbb{R}^+ \rightarrow \mathbb{R}^{n}$ denote the desired position trajectories of $\mathbf{h}_c(\mathbf{q})$, which are generated by the proposed motion planner.
By defining the trajectory tracking errors as $\mathbf{h}(t,\mathbf{q}):=\mathbf{h}_c(\mathbf{q})-\mathbf{h}_d(t)$,
    the control objective becomes to drive $\mathbf{h}$ to zero exponentially.


	
   With the output function $\mathbf{y}$ designed as $\mathbf{h}$, an input-output linearizing control law~\cite{khalil1996noninear} is derived as 
	\begin{equation} 
	\label{equ:feedback control law}
	\mathbf{u=(\tfrac{\partial h}{\partial q} M^{-1}\bar{B})^{-1}[(\tfrac{\partial h}{\partial q} )M^{-1}\bar{c}}
	+\mathbf{v}+\ddot{\mathbf{h}}_d]
	\end{equation}  
	with 
	$$\mathbf{v}= -\mathbf{K}_p \mathbf{y}-\mathbf{K}_d \dot{\mathbf{y}},$$
	where $\mathbf{K}_p \in \mathbb{R}^{n\times n}$
	and
	$\mathbf{K}_d \in \mathbb{R}^{n\times n}$ are both positive definite diagonal matrices.
	
	%
	Then, the continuous-phase closed-loop dynamics in Eq.~\eqref{equ:complete dynamics} become $\mathbf{\ddot{y}}=-\mathbf{K}_d \dot{\mathbf{y}} - \mathbf{K}_p \mathbf{y}$. 
	


	The closed-loop tracking error dynamics can be expressed as:
	\begin{equation} 
    \label{c5-hybrid}
	\begin{cases}
	\dot{\mathbf{x}} 
	= \mathbf{A} \mathbf{x} 
	:=
	\small{
	\begin{bmatrix} 
	\mathbf{0}_{n \times n} & \mathbf{I}_{n \times n} \\
	-\mathbf{K}_p & -\mathbf{K}_d 
	\end{bmatrix}
	}
	\mathbf{x} 
	&\text{if } (t,\mathbf{x^-}) \notin S (t,\mathbf{x});
	\\
	\mathbf{x}^+ 
	= \Delta(t,\mathbf{x}^-)
	&\text{if } (t,\mathbf{x^-}) \in S (t,\mathbf{x}),
	\end{cases}
	\end{equation}
	where
	$\mathbf{x}:=\begin{bmatrix} \mathbf{y}^T,~\dot{\mathbf{y}}^T \end{bmatrix}^T \in \mathcal{X}$ $ \subset \mathbb{R}^{2n}$
	and the expressions of $S: \mathbb{R}^+ \times \mathcal{X} \rightarrow \mathbb{R}^{2n-1} $ and $\Delta: \mathbb{R}^+ \times \mathcal{X} \rightarrow \mathcal{X}$ can be obtained from $S_q$ and $\mathbf{R}_{\dot{q}}$.
	Here, both the reset map and switching surface associated with $\mathbf{x}$ explicitly depend on time because $\mathbf{y}$ is designed as explicitly time-dependent.
	
    
    By the stability conditions based on the construction of multiple Lyapunov functions~\cite{branicky1998multiple}, 
    the closed-loop tracking error dynamics in Eq.~\eqref{c5-hybrid} is locally exponentially stable if there exists a Lyapunov function candidate $V(\mathbf{x})$ and a positive number $r$ such that a) $V(\mathbf{x})$ exponentially decreases during each continuous phase and b) $\{V |^+_{1}, V|^+_{2}, V|^+_{3} ... \}$ is monotonically decreasing for any $\mathbf{x}(T_0) \in B_{r} (\mathbf{0}) :=\{ \mathbf{x}:   \| \mathbf{x} \| \leq r  \}$.
    Here, $V(T_{k}^+):=V|^+_{k}$ ($k \in \{1,2,... \}$). $T_K$ is the $K^{th}$ actual impact time, and $\tau_K$ is its desired value provided by the proposed planner.
    
    With properly chosen $\mathbf{K}_p$ and $\mathbf{K}_d$, the tracking error $\mathbf{y}$ will exponentially diminishes during continuous phases, which indicates that the condition a) is met.
    
    To guarantee the condition b) is met for ensuring the closed-loop stability, it is necessary to analyze the rest map in Eq.~\eqref{c5-hybrid}.
    From Eq.~\eqref{c5-hybrid}, one has
\begin{equation}
\begin{aligned}
\| \mathbf{x}|^+_{K} \| 
=& \| \Delta (T_K^-, \mathbf{x}|^-_K )  \| 
\\
\leq
& 
\| \Delta (T_{K}^-, \mathbf{x}|^-_{K} )  -  \Delta (\tau_{K}^-, \mathbf{x}|^-_{K}  )     \| 
\\
&+ 
\| \Delta (\tau_{K}^-, \mathbf{x}|^-_{K}  )  -  \Delta (\tau_{K}^-, \mathbf{0} )     \| \\
&+
\| \Delta (\tau_{K}^-, \mathbf{0} )     \|.
\end{aligned}
\label{c5-eq49}
\end{equation} 

If the desired trajectories $\mathbf{h}_d(t)$ is generated to be smooth, then the reset map $\Delta$ will be continuously differentiable in $t$.
Also, $\Delta$ is continuously differentiable in $\mathbf{x}$~\cite{grizzle2001asymptotically}.
Therefore, there exist positive numbers $r_1$, $L_{\Delta_t}$, and $L_{\Delta_x}$ such that 
$\| \Delta (T_{K}^-, \mathbf{x}|^-_{K}  )  
-
\Delta (\tau_{K}^-, \mathbf{x}|^-_{K}  )     \| 
\leq 
L_{\Delta_t} \| T_{k} - \tau_{k}  \|$ holds for any $\mathbf{x}(T_0) \in B_{r_1}(\mathbf{0})$~\cite{gu2016bipedal},
along with $\| \Delta (\tau_{K}^-, \mathbf{x}|^-_{K}  )  -  \Delta (\tau_{K}^-, \mathbf{0}  )     \|  \leq L_{\Delta_x} \| \mathbf{x}|^-_{K} \|$.

If the desired motion is planned as impact-aware, then based on our previous analysis~\cite{gu2016bipedal,yuan_2018}, it can be proved that
the convergence rate of the sequence $ \{ \mathbf {x}|^+_{1}, \mathbf {x}|^+_{2}, ... \} $ can be directly tuned by the PD control gains $K_p$ and $K_d$~\cite{yuan_2018}.
Therefore, the sequence $\{V |^+_{1}, V|^+_{2}, V|^+_{3} ... \}$ will be monotonically decreasing with properly chosen PD control gains.

\section{SIMULATIONS}
	
	\label{Section-Simulation}
This section presents the MATLAB~\cite{Matlab_link} and Webots~\cite{Webots_link} simulation results for demonstrating the effectiveness of our proposed online planning strategy in generating impact-aware, dynamically feasible full-body motion. 
MATLAB simulations were performed for initial validations, whereas Webots simulations were intended for more realistic validations, which can be used to guide future experimental validation.
For the convenience of comparison, both MATLAB and Webots simulations use the same footstep sequences (with timestamps) $\Gamma$ and the same number of sample points $K=6$ as the input to the proposed planner.

Overall, our planner takes within 0.6 second to generate one walking step of impact-aware full-body motion, which is shorter than a typical bipedal walking step.
As discussed in the Section~\ref{Section-planner}, the complete planning task is decomposed into four subtasks.
The subtasks of posture interpolation and computing keyframe postures take approximately $0.05$ second to compute in total.
The subtask of pre- and post-impact velocity assignment takes approximately 0.13 second to compute.
The subtask of continuous-phase motion generation takes approximately 0.4 second to compute.
While the nonlinear optimization problem associated with the subtask of continuous-phase motion generation is highly sparse,
it can be solved efficiently using IPOPT~\cite{Wachter2006} solver in the optimization framework CasADi~\cite{andersson2019casadi}, thus resulting in a short planning time of approximately 0.4 second.


\subsection{Trajectory Interpolation}

As the output of the planner comprises isolated way-points, trajectory interpolation is needed to generate the desired continuous trajectory.
Thus, before presenting the simulation results of the proposed planner, common trajectory interpolation techniques are briefly discussed next.

  Piecewise cubic Hermite functions are commonly used for trajectory interpolation, including: Piecewise Cubic Hermite Interpolating Polynomial (PCHIP), Cubic Spline Data Interpolation (SPLINE), and Modified Akima Piecewise Cubic Hermite Interpolation (MAKIMA).

  
  The SPLINE function produces the smoothest trajectory amongt the three. 
  However, this method suffers from large overshoot, resulting in severe distortion in the interpolated motion.
  The PCHIP function does not suffer the overshoot issue and thus preserves the shape of the interpolated trajectories.
  However, the smoothness of the interpolated trajectories will not be preserved.  
  The performance of the MAKIMA function lies between the PCHIP function and the SPLINE function. 
  In this work, we choose to use PCHIP to interpolate continuous trajectories for preserving the shape of the interpolated trajectories and avoiding motion distortion.
  Although the controller may suffer from large control effort at the non-smooth points, this can be mitigated by adjusting control gains.
  As the focus of this study is on online impact-aware motion planning, the trajectory interpolation technique used here is not intended to be optimal, which will be further addressed in future investigations.

\subsection{MATLAB Simulation}

In MATLAB simulation, the input-output linearizing control strategy as introduced in Section \ref{Section-Control} is applied to drive the robot to the planner motion.
The dynamic matrices, such as $\mathbf{M(q)}$ and $\mathbf{c(q,\dot{q})}$, can be computed efficiently using FROST~\cite{hereid2017frost}.

\begin{figure}[h]
    \centering
    \includegraphics[width=1\linewidth]{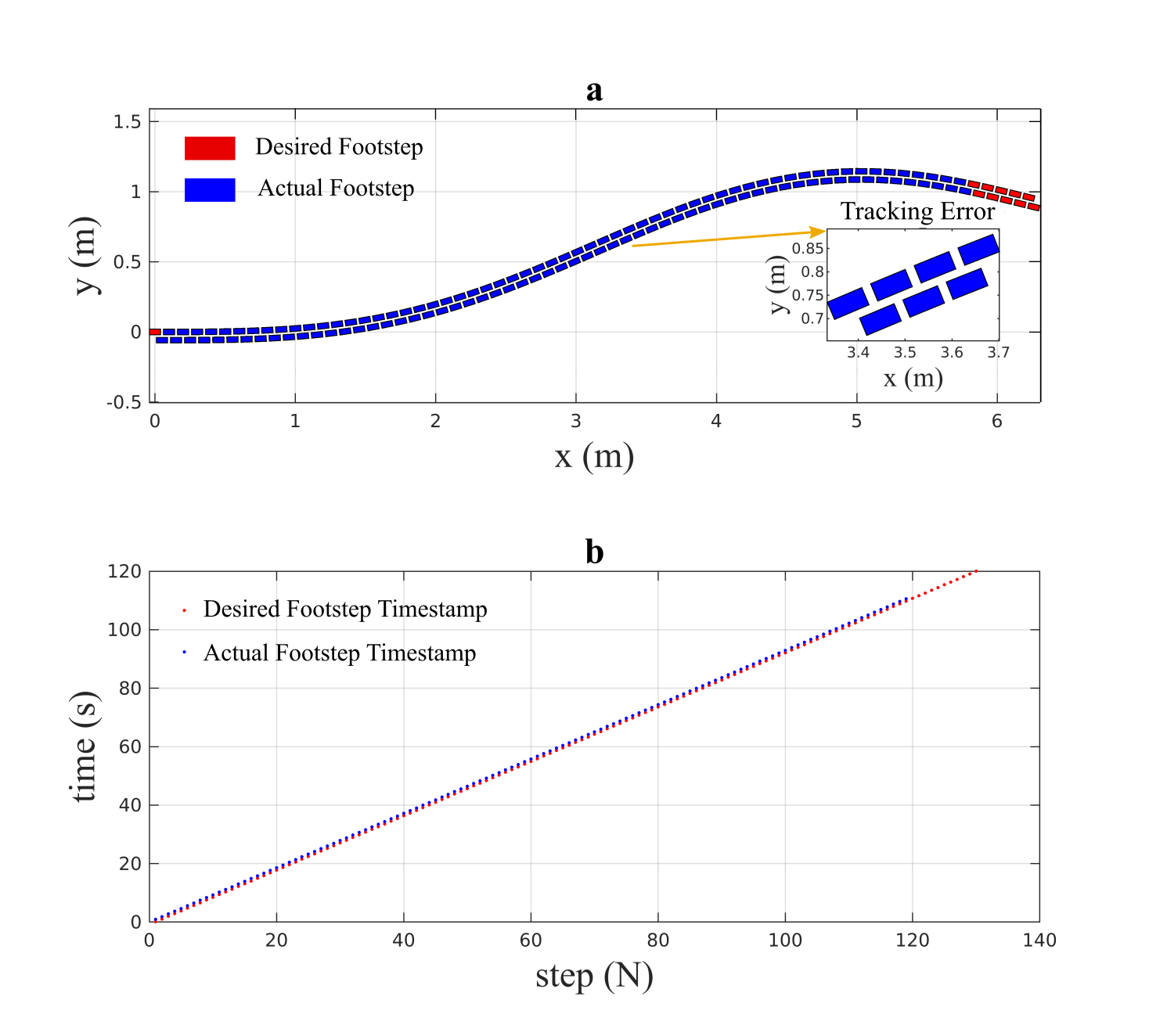}
    \caption{MATLAB simulation results of a) satisfactory footstep tracking and b) satisfactory convergence of foot-landing timing.
    130 steps are tracked in total.}
    \label{fig:MAT footstep tracking}
\end{figure}

From the simulation results (Fig.~\ref{fig:MAT footstep tracking}), it is clear that our planning and control strategies result in satisfactory tracking of the desired footsteps with specific timing.
The results demonstrate the effectiveness of our proposed planning strategy in generating impact-aware, dynamically feasible trajectory in real-time.

\subsection{Webots Simulation}

\begin{figure}[h]
    \centering
    \includegraphics[width=1\linewidth]{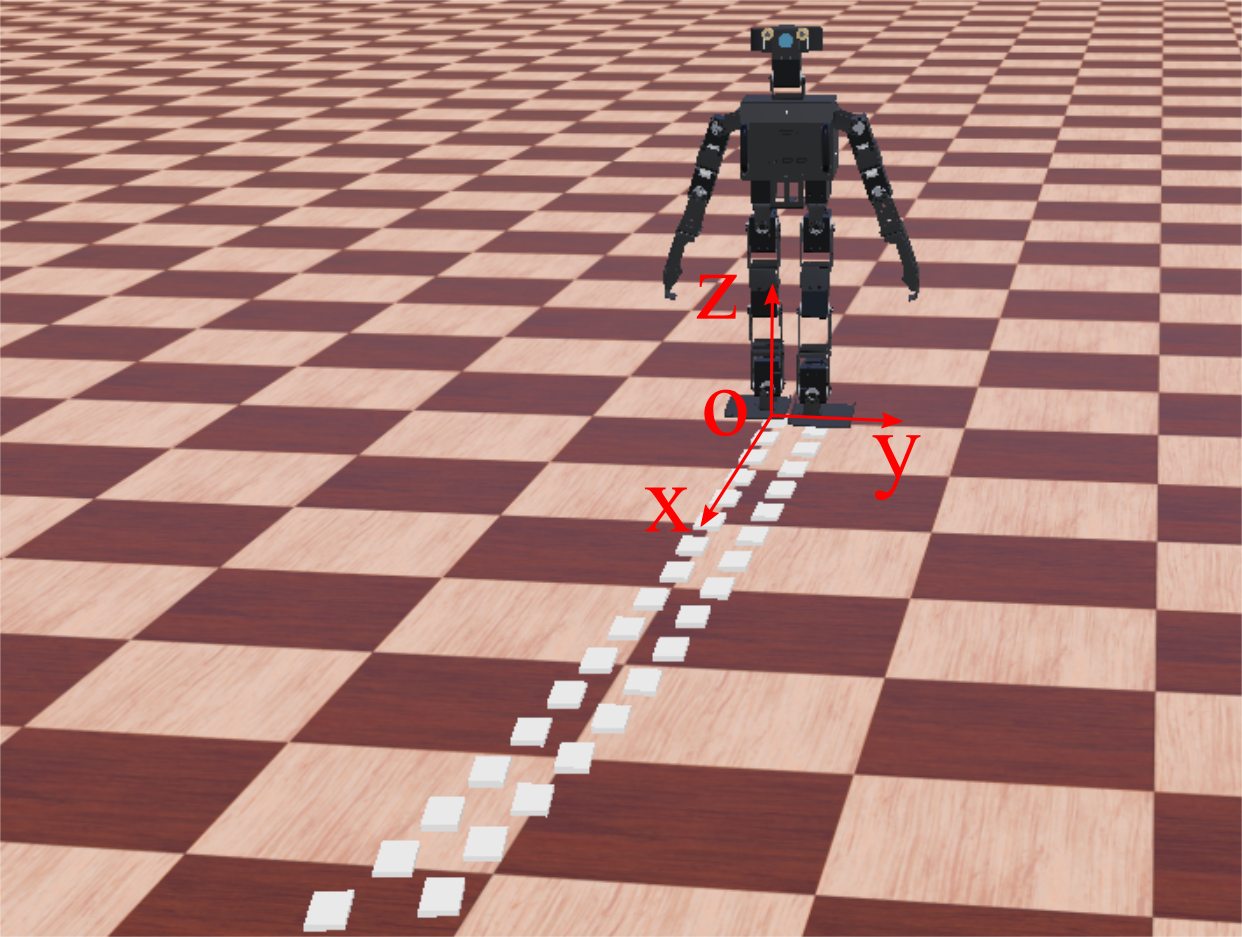}
    \caption{Simulation setup in Webots. The shaded rectangles indicate the desired footstep sequence.}
    \label{fig:webots setup}
\end{figure}

\begin{figure}[h]
    \centering
    \includegraphics[width=1.0\linewidth]{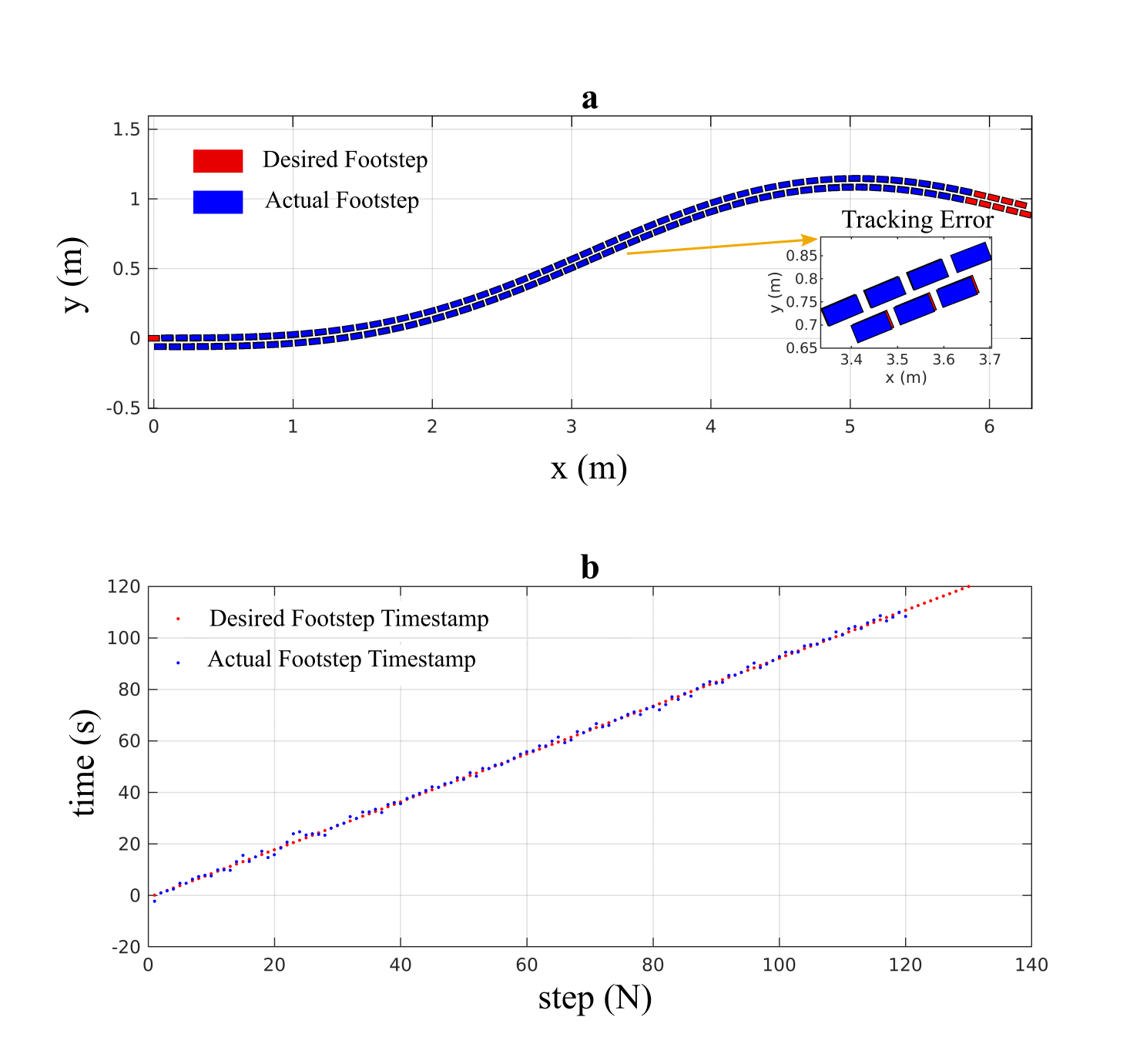}
    \caption{Webots simulation results of a) satisfactory footstep tracking and b) satisfactory convergence of foot-landing timing.
    130 steps are tracked in total.}
    \label{fig:webots footstep tracking}
\end{figure}

In Webots simulation, we use the same optimization framework as implemented in MATLAB to generate the desired motion online. 
The setup of the Webots is illustrated in Fig.~\ref{fig:webots setup}. 
For simplicity and without generality, individual joint control adapted from the input-output linearizing control~\cite{yuan_2018} is utilized in the Webots simulation.
Fig.~\ref{fig:webots footstep tracking} shows the footstep tracking results in Webots.
As compared with the MATLAB results, the tracking performance in Webots is less accurate because the individual joint controller ignores the nonlinear coupling among joints.
However, as the steady-state tracking error is small and bounded, we can still consider that Webots simulation demonstrated a reasonably good trajectory tracking performance in terms of footstep tracking with specific timing.

\section{CONCLUSIONS}

    In this paper, we have introduced an online planning method that generates impact-aware, dynamically feasible, full-body desired trajectories for fully actuated bipedal walking robots. 
    There are four main components of the proposed planner, including
    \YuanMod{posture interpolation, computing keyframe posture, keyframe posture velocity assignment, and full-body motion generation based on reduced-order dynamics and full-order kinematics}.
    To validate the proposed planner through simulations on a fully actuated bipedal walking robot, a provable trajectory tracking control law was synthesized and simulated to track the generated motions.
    Results of both MATLAB and 3-D realistic simulations demonstrated the effectiveness of the proposed online planning strategy in generating dynamically feasible, full-body motions that respect both the discrete dynamics and desired timing of the given footstep sequence.

\label{section-conclusions}
\balance
\bibliography{Reference1}

\bibliographystyle{ieeetr}

\end{document}